\documentclass[conference]{IEEEtran}
\IEEEoverridecommandlockouts

\usepackage{hyperref}
\usepackage{float}
\usepackage{soul}
\usepackage{amsmath,amssymb,amsfonts}
\usepackage{scalerel}
\usepackage{algorithmic}
\usepackage{graphicx}
\usepackage{textcomp}
\usepackage{adjustbox}
\usepackage{xcolor}
\def\BibTeX{{\rm B\kern-.05em{\sc i\kern-.025em b}\kern-.08em
    T\kern-.1667em\lower.7ex\hbox{E}\kern-.125emX}}
\usepackage{caption}
\usepackage{subcaption}
\usepackage[backend=biber,style=ieee]{biblatex}
\AtEveryBibitem{\clearfield{doi}
                \clearfield{note}
                \clearfield{urldate}
                \clearfield{issn}
                \clearfield{isbn}
                \clearfield{language}
                }
\AtEveryBibitem{%
  \iffieldequalstr{entrykey}{lecun_mnist_1998}
    {}
    {\clearfield{url}}%
}
\DeclareSourcemap{
  \maps[datatype=bibtex]{
    \map{
      \step[fieldset=urldate, null]
    }
  }
}

\newcommand{\figurewidth}{0.44}

\begin{document}

\title{A Method for Optimizing Connections in Differentiable Logic Gate Networks
\thanks{Flanders AI Research}
}

\author{\IEEEauthorblockN{Wout Mommen\IEEEauthorrefmark{1}\IEEEauthorrefmark{2}, Lars Keuninckx\IEEEauthorrefmark{1}, Matthias Hartmann\IEEEauthorrefmark{1}, Piet Wambacq\IEEEauthorrefmark{2}}\\
\IEEEauthorblockA{\IEEEauthorrefmark{1}Novel Algorithms and Compute Systems (NACS)\\
imec, Leuven, Belgium\\
Email: Wout.Mommen@imec.be}\\

\IEEEauthorblockA{\IEEEauthorrefmark{2} Electronics and Informatics (ETRO)\\
Vrije Universiteit Brussel (VUB), Brussels, Belgium\\
Email: Piet.Wambacq@vub.be}}

\maketitle

\begin{abstract}
We introduce a novel method for partial optimization of the connections in Deep Differentiable Logic Gate Networks (LGNs). Our training method utilizes a probability distribution over a subset of connections per gate input, selecting the connection with highest merit, after which the gate-types are selected. We show that the connection-optimized LGNs outperform standard fixed-connection LGNs on the Yin-Yang, MNIST and Fashion-MNIST benchmarks, while requiring only a fraction of the number of logic gates. When training all connections, we demonstrate that 8000 simple logic gates are sufficient to achieve over 98\% on the MNIST data set. Additionally, we show that our network has 24 times fewer gates, while performing better on the MNIST data set compared to standard fully connected LGNs. As such, our work shows a pathway towards fully trainable Boolean logic.\\
\end{abstract}


\section{Introduction}
In recent years it has become clear that running AI models requires a lot of power \cite{sevilla_compute_2022, desislavov_trends_2023}. As such, a strong interest has been shown to make these models and their associated hardware more energy efficient, especially on the inference side \cite{rastegari_xnor-net_2016,hubara_quantized_2018}. One example is the field of neuromorphic computing that uses bio-inspired Spiking Neural Networks (SNNs) \cite{schuman_opportunities_2022,eshraghian_training_2023} to save a great deal of energy by considering the inputs to all layers to be binary. In that way, the multiply-accumulate (MAC) operations that are needed for determining the neuron activations are reduced to accumulate-only operations. These SNNs can be sparsified \cite{rathi_stdp-based_2019} and their weight values quantized \cite{putra_q-spinn_2021} to a low number of bits, again saving energy and using less hardware.  Clearly, neuromorphic methods such as SNNs have delivered their successes in energy efficiency \cite{diehl_truehappiness_2016,kim_spiking-yolo_2020}, however the question arises if this is the best solution possible.\\

Very recently, it has become clear that directly training a network of Boolean gates, i.e. the very fabric of digital hardware, is possible \cite{petersen_deep_2022}. These so-called Deep Differentiable Logic Gate Networks (LGNs) are feedforward multi-layered networks of logic gates, in which each gate has two input connections. These networks have fixed random connections but trainable gate-types, which are relaxed to 16 different differentiable expressions, one for each two-input Boolean operator, and are trained with backpropagation. A convolutional approach to LGNs has also been demonstrated showing state-of-the-art performance in throughput speed, accuracy and number of gates \cite{petersen_convolutional_2024}. Additionally, there exist methods that use other digital hardware elements as ``neurons'' such as monostable multivibrator (MMV) networks \cite{keuninckx_not_2023}, lookup table (LUT) networks \cite{wang_lutnet_2019,wang_lutnet_2020} and networks of multiplexers (MUXes) \cite{lee_stochastic_2024}.\\

In this work we propose a method to extend the paradigm of LGNs with trainable connections.  
Our contributions are the following:
\begin{itemize}
    \item We introduce a novel method for learning the connections in LGNs. Currently, the connections in LGNs are fixed, hence it is likely not the optimal way of obtaining the smallest network of logic gates for a certain task. In this work, the number of learnable connections per gate input is a hyperparameter that can be chosen. 
    \item This method is tested on the Yin-Yang, MNIST and Fashion-MNIST benchmarks. It is shown that LGNs with trained connections are significantly smaller compared to LGNs with fixed connections, while maintaining the same state-of-the-art accuracy and throughput.
    \item A first attempt at training all connections in LGNs with this method is tried out. Although the increase of accuracy is only valid up to two layer deep networks, we show that our fully connected network needs 24 times fewer gates compared to fixed connection networks, while achieving a higher accuracy on the MNIST data set. Even when comparing our fully connected network to a convolutional network with fixed connections, we still use 9 times less gates.
\end{itemize}

\section{Training method}
\subsection{Standard training method}
The standard way of training the gates in LGNs is by learning a probability distribution $(p_g)_{k,i}^l$ over all 16 possible 2-input Boolean operations $f_i$ (e.g. AND, OR, XNOR, ...). Here $l$ is the layer number, $k$ is the index of the gate within that layer and $i$ refers to gate operation $f_i$. Each of these 16 operations is explained in \cite{petersen_deep_2022}. The output of gate $k$ in layer $l$ is represented in the following way:

\begin{equation} \label{eq:softmax_G}
g_k^l=\sum_{i=0}^{15} (p_g)_{k,i}^l \cdot f_i\left(a_k^{l}, b_k^{l}\right)=\sum_{i=0}^{15} \frac{e^{(w_g)_{k,i}^l/T_g}}{\sum_j e^{(w_g)_{k,j}^l/T_g}} \cdot f_i\left(a_k^l, b_k^l\right).
\end{equation}

Here $a^{l}_k$ and $b^{l}_k$ are the input values to logic gate $k$ of layer $l$. The probability distribution $(p_g)_{k,i}^l$ is given by the output of a softmax function that contains trainable weights $(w_g)_{k,i}^l$, such that for each gate  in each layer there is an associated weight for each gate operation $f_i$. The temperature $T_g$ is a hyperparameter. After training the gates in a network with non-trainable connections, the gate operation $f_i$ matching the largest probability is chosen as in \cite{petersen_deep_2022}. Residual initialization is used to initialize weight values of the gates as described in \cite{petersen_convolutional_2024}.\\

\subsection{Training the connections}
We now extend this method of learning the correct gate operation for each gate through a softmax probability distribution, to learning the connections to the gates. The first input $a$ of gate $k$ in layer $l$ is given by\\
\begin{equation} \label{eq:softmax_C}
a_{k}^{l}=\sum_{i=0}^{N_c-1} (p_a)_{k,i}^{l} \cdot g_{i}^{l-1}=\sum_{i=0}^{N_c-1} \frac{e^{(w_a)_{k,i}^l/T_c}}{\sum_j e^{(w_a)_{k,j}^l/T_c}} \cdot g_{i}^{l-1}.    
\end{equation}

Here $N_c$ are the number of connections that are trained for each input of each gate, which is equal to $N^{(l-1)}$ when all connections are trained. A weight matrix $(w_a)_{k,i}^l$ for the first input of gate $k$ in layer $l$ is learned and each of the weight values are normalized by a connection temperature $T_c$. The index $i$ refers to the $i^{th}$ gate in the previous layer $(l-1)$. The expression for the second input $b_k^l$ is similar, but with probability $(p_b)_{k,i}^{l}$ and weight matrix $(w_b)_{k,i}^l$. Similarly to the gates, the softmax is changed into an argmax distribution, selecting the connection with the highest probability. This is done by starting with a high $T_c$ and slowly lowering it to a chosen minimum value during training. In that way, the output of the softmax function changes from having a very uniform distribution at the beginning of training, to a very peaked one at the end of training. Hence, the softmax function can be seen as a smooth argmax function. Implementation-wise, if $N_c = N^{(l-1)}$, a weight matrix is used that contains full precision values at the start of training. A softmax is calculated for each row of the $(w_a)_{k,i}^l$ and $(w_b)_{k,i}^l$ matrices. A matrix-vector multiplication between these matrices and the inputs are performed. By slowly lowering $T_c$, each row will converge to an $N_c$-dimensional one-hot encoded vector. If $N_c$ is quite small (e.g. 8 or 16), only indices are used to refer to inputs of the previous layer, which requires less memory. The matrix-vector multiplication is then replaced by the Hadamard product. In the next section, we apply our method to three different common data sets.

\section{Results and discussion}
\subsection{Yin-Yang}
\begin{figure}[t]
    \centering
     \includegraphics[width=0.4\textwidth,keepaspectratio]{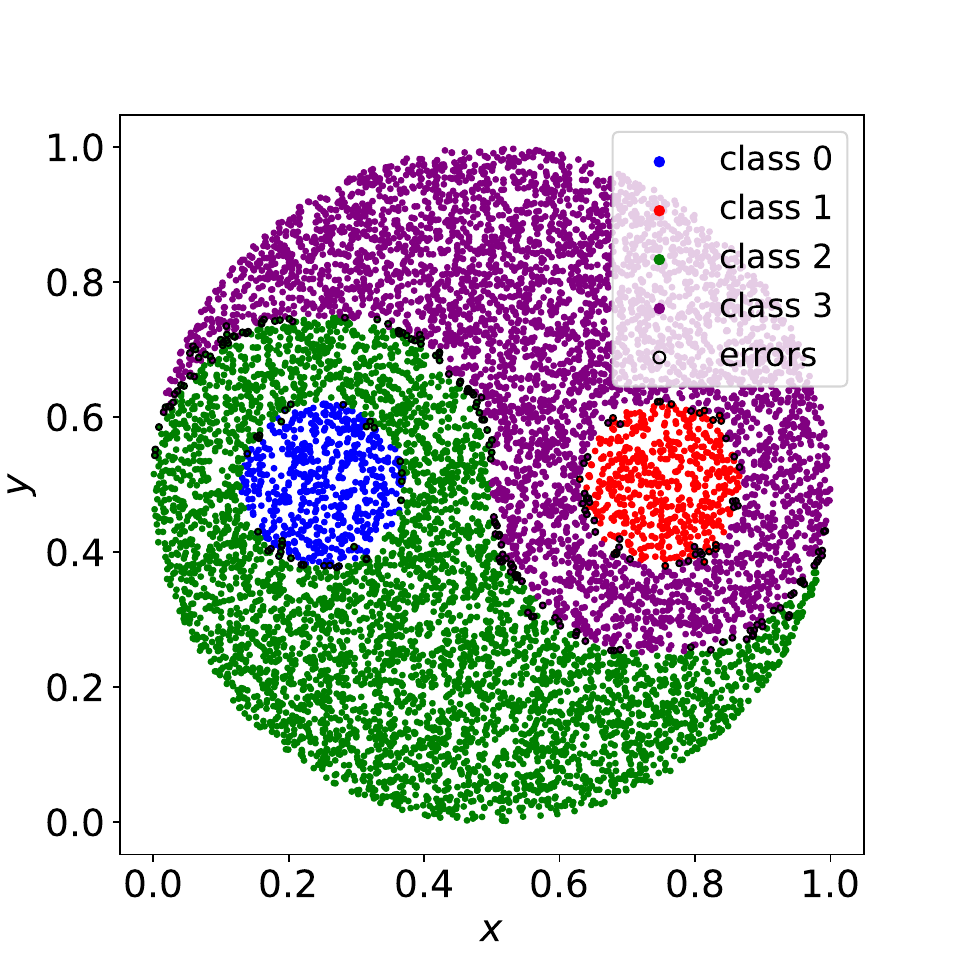}
     \caption{The test output of the Yin-Yang data set using a 2-layer network of 500 gates/layer with the number of trainable connections per gate input ($N_c$) equal to 16. A test accuracy of 99.22\% was observed on these 10 000 data points.}
     \label{fig:YinYang}
\end{figure}

\begin{figure}[t]
     \centering
     \begin{subfigure}[]{\figurewidth\textwidth}
         \centering
         \includegraphics[width=\textwidth]{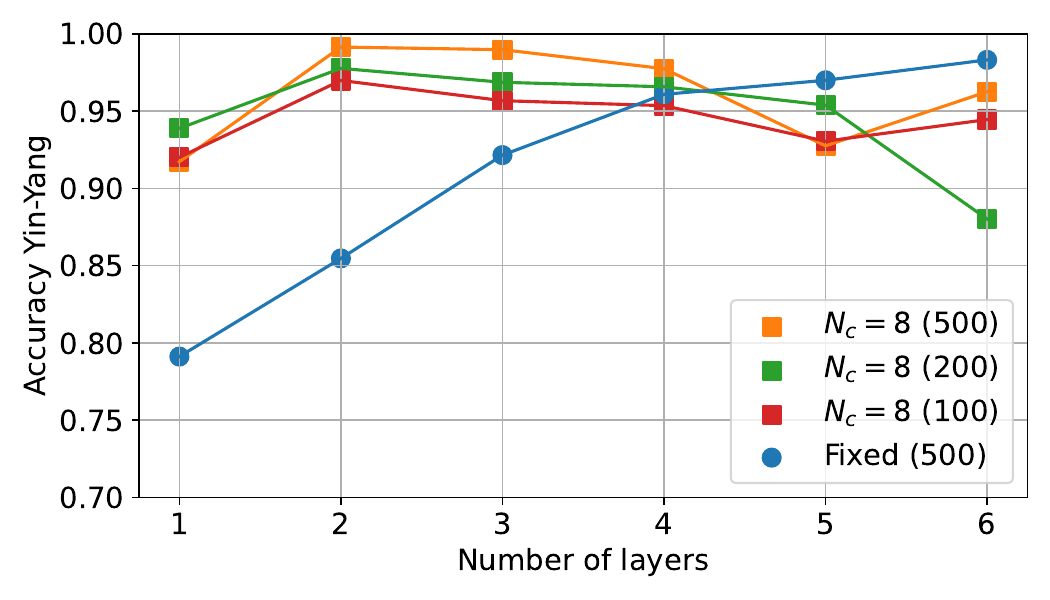}
         \caption{}
         \label{fig:YinYang_Nc8}
     \end{subfigure}
     \hfill
     \begin{subfigure}[]{\figurewidth\textwidth}
         \centering
         \includegraphics[width=\textwidth]{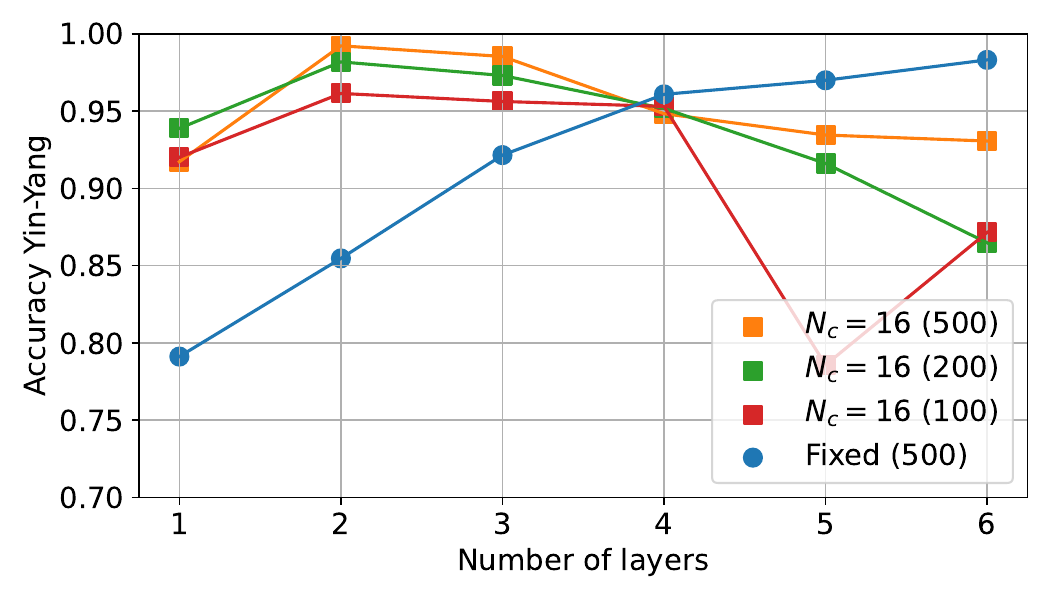}
         \caption{}
         \label{fig:YinYang_Nc16}
     \end{subfigure}
        \caption{Accuracy on the Yin-Yang data set as a function of number of layers for (a) $N_c=8$ and (b) $N_c = 16$ compared with a network of fixed connections. Here $N_c$ is the number of trainable connections per input per gate. The number of gates per layer are given between brackets.}
        \label{fig:YinYang_training}
\end{figure}

The goal of the Yin-Yang data set \cite{kriener_yin-yang_2022} is to classify points in the $xy$-plane into four different regions, as given in Figure\,\ref{fig:YinYang}. The $(x,y)$ coordinates are encoded in 12-bit vectors, such that the input dimension of the networks is 24. The train set consists of 200\,000 data points and the test set contains 10\,000 data points. The accuracy as a function of layer depth can be seen in Figure \ref{fig:YinYang_Nc8} for $N_c=8$ and Figure \ref{fig:YinYang_Nc16} for $N_c=16$. A value of 12 is chosen for $N_c$ for the first layer. All models are trained with a learning rate of 0.01 for 100 epochs. For the models with trainable connections, $T_c$ is lowered from 1 tot $10^{-4}$ from epoch 60 to 80 and $T_g$ is lowered from 1 tot $10^{-4}$ from epoch 80 to 100. From Figure \ref{fig:YinYang_Nc8} and Figure \ref{fig:YinYang_Nc16} it is clear that training the connections improves the accuracy significantly for shallow networks, since all models with trainable connections perform better up to around 3-4 layers. For example, a network with trainable connections consisting of 200 gates (2 layers of 100 gates/layer) achieves a higher accuracy (96.98\%) than a network with fixed connections of 2000 gates (4 layers of 500 gates/layer, 96.08\%), resulting in one order of magnitude less hardware. Due to a smaller output layer size of 100 gates/layer compared to 500 gates/layer, the population counters will be smaller, again saving hardware.

\subsection{MNIST}
\begin{figure}[t]
     \centering
     \begin{subfigure}[h!]{\figurewidth\textwidth}
         \centering
         \includegraphics[width=\textwidth]{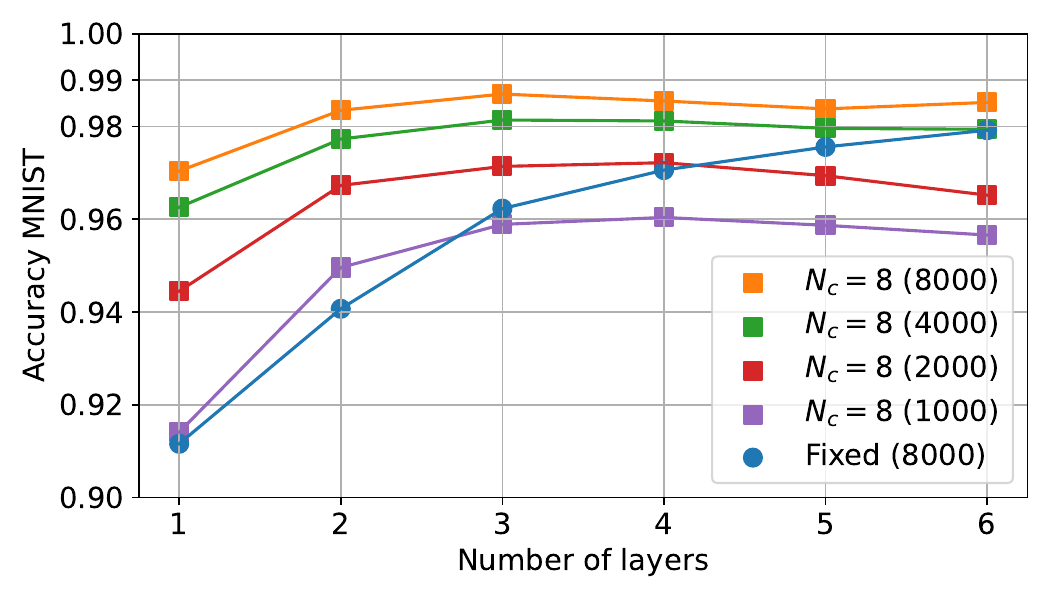}
         \caption{}
         \label{fig:MNIST_Nc8}
     \end{subfigure}
     \hfill
     \begin{subfigure}[h!]{\figurewidth\textwidth}
         \centering
         \includegraphics[width=\textwidth]{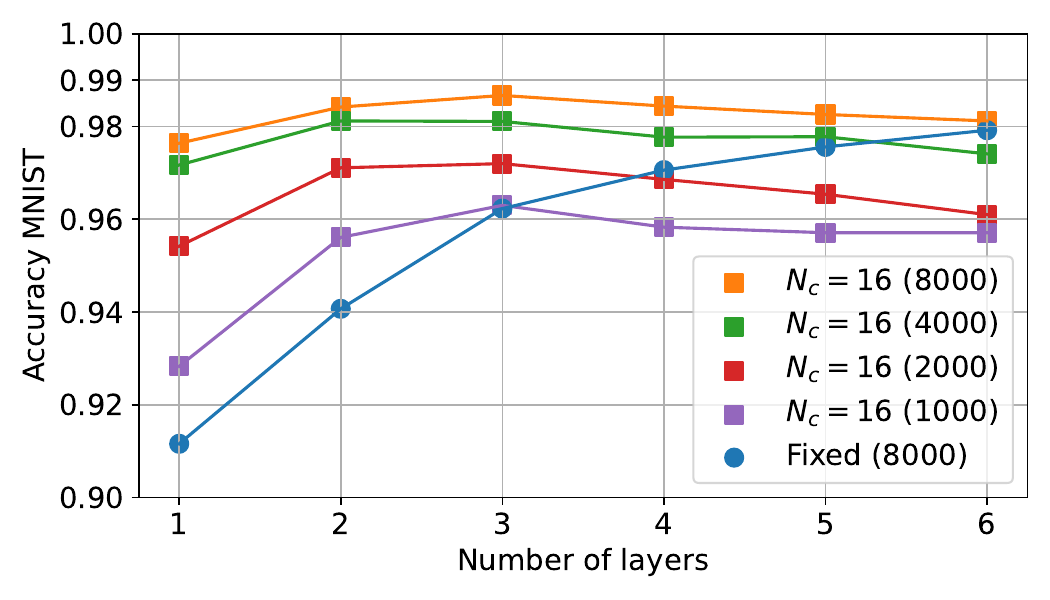}
         \caption{}
         \label{fig:MNIST_Nc16}
     \end{subfigure}
        \caption{Accuracy on the MNIST data set as a function of number of layers for (a) $N_c=8$ and (b) $N_c = 16$ compared with a network of fixed connections. Here $N_c$ is the number of trainable connections per input per gate. The number of gates per layer are given between brackets.}
        \label{fig:MNIST_training}
\end{figure}

The MNIST data set \cite{lecun_mnist_1998} consist of 28x28 grayscale images of handwritten digits ranging from 0 until 9. Before inputting these 784 pixel values into the model, a threshold of 0.5 is used to binarize them. Figure \ref{fig:MNIST_Nc8} and \ref{fig:MNIST_Nc16} give the accuracy of the MNIST data set as a function of layer depth for $N_c=8$ and $N_c=16$ respectively. All models are trained using a learning rate of 0.01. The models with fixed or trainable connections are trained for 200 or 240 epochs respectively. In the latter case, $T_c$ is lowered from epoch 160 until 200 and $T_g$ is lowered from epoch 200 until 240, using the same start and end values as mentioned before. Again, the networks with trainable connections outperform the ones with fixed connections, using the same number of gates/layer. The largest improvement is found for more shallow networks, i.e. a 3-layer network of 4000 gates/layer with trainable connections performs better (98.14\%) than a 6-layer network of 8000 gates/layer (97.92\%). This is an improvement by a factor of 4 with respect to the number of gates and connections. Another example is the 1-layer network of 1000 gates with $N_c=16$, which performs better (92.83\%) than a 1-layer network with fixed connections (91.16\%), while having 8 times less gates and connections. Again, it is expected that less logic is needed for the population counts.

\subsection{Fashion-MNIST}
\begin{figure}[t]
     \centering
     \begin{subfigure}[h!]{\figurewidth\textwidth}
         \centering
         \includegraphics[width=\textwidth]{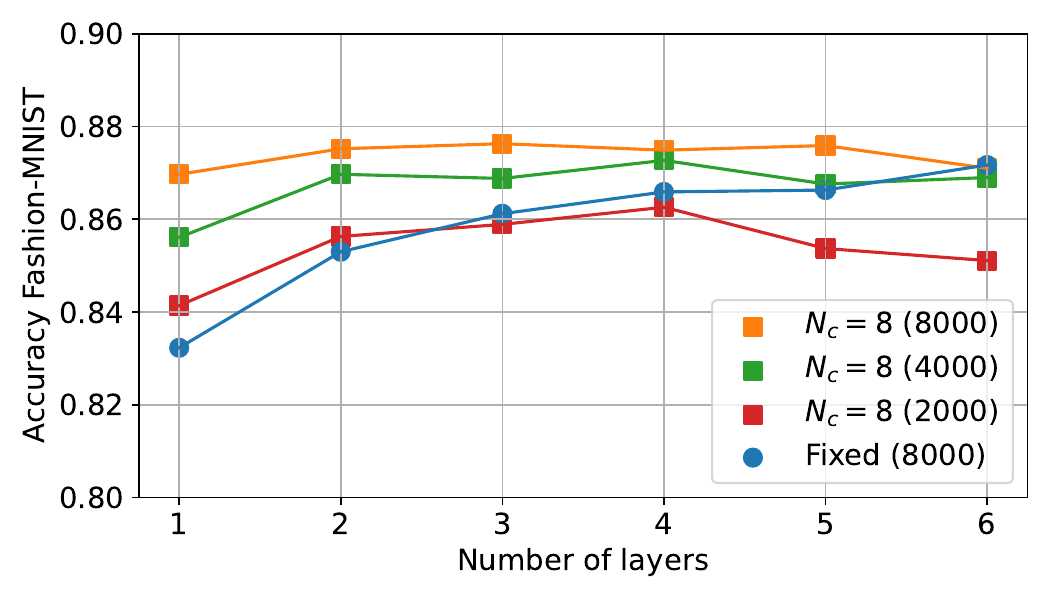}
         \caption{}
         \label{fig:FashionMNIST_Nc8}
     \end{subfigure}
     \hfill
     \begin{subfigure}[h!]{\figurewidth\textwidth}
         \centering
         \includegraphics[width=\textwidth]{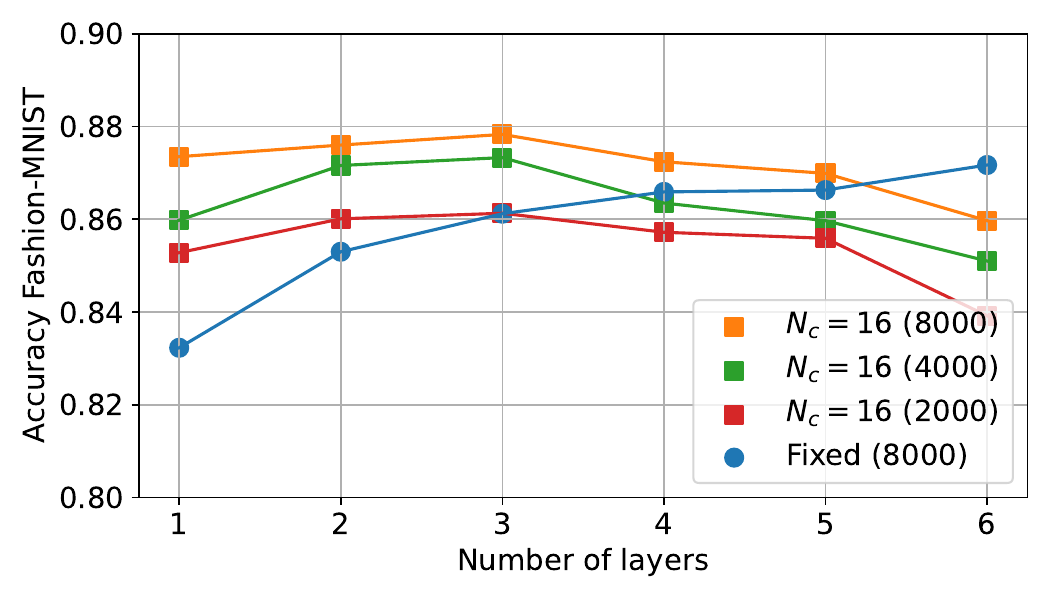}
         \caption{}
         \label{fig:FashionMNIST_Nc16}
     \end{subfigure}
        \caption{Accuracy on the Fashion-MNIST data set as a function of number of layers for (a) $N_c=8$ and (b) $N_c = 16$ compared with a network of fixed connections. Here $N_c$ is the number of trainable connections per input per gate. The number of gates per layer are given between brackets.}
        \label{fig:FashionMNIST_training}
\end{figure}

The Fashion-MNIST data set \cite{xiao_fashion-mnist_2017} is similar to the MNIST data set, but instead of handwritten digits, there are 10 classes of clothing items. Thresholds of 0.25, 0.5 and 0.75 are used to encoded the data into a binary format such that it can be used as input to the models. All models are trained using the same settings as for the MNIST data set. The accuracy as a function of number of layers can be seen in Figure \ref{fig:FashionMNIST_Nc8} and Figure \ref{fig:FashionMNIST_Nc16} for $N_c = 8$ and $N_c=16$ respectively. Up to 5 layers, the networks with trainable connections clearly outperform the networks with fixed connections. For example, a network with trainable connections using 2 layers of 4000 gates/layer has a similar accuracy (87.16\%) as a network using 6 layers of 8000 gates/layer (87.17\%) with fixed connections, requiring 6 times less gates and connections.

\subsection{Training all connections}
We attempt to train all connections of LGNs on the MNIST data set. An accuracy increase up to two layer deep networks was achieved, but for deeper networks the accuracy slightly decreased. Using a network of 1x8000 and 2x8000 gates, accuracies of 98.19\% and 98.68\% were observed respectively, showcasing that 8000 simple logic gates are all you need to obtain more than 98\% on MNIST. According to \cite{petersen_convolutional_2024}, 384\,000 gates (fully connected) and 147 000 gates (convolutional) are needed to obtain accuracies of 98.47\% and 98.46\% respectively. This means that our fully connected network achieves a higher accuracy (98.68\%), while having a factor of 24 (fully connected) and 9 (convolutional) fewer gates.

\section{Conclusion}
Partially training the connections in LGNs leads to a better performance using significantly smaller networks, especially for shallow networks. It was observed that the networks trained on the Yin-Yang, MNIST and Fashion-MNIST data sets needed 10, 8 and 6 times fewer gates and connections respectively, while having an equivalent or higher accuracy. When training all connections of a fully connected network, this factor was equal to 24 (fully connected with fixed connections) or 9 (convolutional with fixed connections). In future work we will focus on further improving the algorithm to deeper networks and applying it to Convolutional Differentiable LGNs. Only training a low number of connections can already be useful for fully training the convolutional filters in these types of networks. The classification part is usually quite shallow and could be trained fully. Ultimately, this leads to more efficient scaling for larger networks, broadening the range of applications of LGNs.

\section*{Acknowledgment}
This work has received funding from the Flemish Government under ``Onderzoeksprogramma Artificiële Intelligentie Vlaanderen'' (Flanders AI Research).
\printbibliography

@inproceedings{diehl_truehappiness_2016,
	title = {{TrueHappiness}: {Neuromorphic} emotion recognition on {TrueNorth}},
	shorttitle = {{TrueHappiness}},
	url = {https://ieeexplore.ieee.org/abstract/document/7727758},
	doi = {10.1109/IJCNN.2016.7727758},
	urldate = {2025-04-06},
	booktitle = {2016 {International} {Joint} {Conference} on {Neural} {Networks} ({IJCNN})},
	author = {Diehl, Peter U. and Pedroni, Bruno U. and Cassidy, Andrew and Merolla, Paul and Neftci, Emre and Zarrella, Guido},
	month = jul,
	year = {2016},
	note = {ISSN: 2161-4407},
	pages = {4278--4285},
}

@article{kim_spiking-yolo_2020,
	title = {Spiking-{YOLO}: {Spiking} {Neural} {Network} for {Energy}-{Efficient} {Object} {Detection}},
	volume = {34},
	copyright = {Copyright (c) 2020 Association for the Advancement of Artificial Intelligence},
	issn = {2374-3468},
	shorttitle = {Spiking-{YOLO}},
	url = {https://ojs.aaai.org/index.php/AAAI/article/view/6787},
	doi = {10.1609/aaai.v34i07.6787},
	language = {en},
	number = {07},
	urldate = {2025-04-06},
	journal = {Proceedings of the AAAI Conference on Artificial Intelligence},
	author = {Kim, Seijoon and Park, Seongsik and Na, Byunggook and Yoon, Sungroh},
	month = apr,
	year = {2020},
	note = {Number: 07},
	pages = {11270--11277},
}

@misc{lecun_mnist_1998,
	title = {{MNIST} database of handwritten digits},
	url = {http://yann.lecun.com/exdb/mnist/},
	urldate = {2025-04-04},
	author = {LeCun, Yann and Cortes, Corinna and J.C. Burges, Christopher},
	year = {1998},
	note = {keepurl: true},
}

@misc{xiao_fashion-mnist_2017,
	title = {Fashion-{MNIST}: a {Novel} {Image} {Dataset} for {Benchmarking} {Machine} {Learning} {Algorithms}},
	shorttitle = {Fashion-{MNIST}},
	url = {http://arxiv.org/abs/1708.07747},
	doi = {10.48550/arXiv.1708.07747},
	urldate = {2025-04-04},
	publisher = {arXiv},
	author = {Xiao, Han and Rasul, Kashif and Vollgraf, Roland},
	month = sep,
	year = {2017},
	note = {arXiv:1708.07747 [cs]},
}

@misc{kriener_yin-yang_2022,
	title = {The {Yin}-{Yang} dataset},
	url = {http://arxiv.org/abs/2102.08211},
	doi = {10.48550/arXiv.2102.08211},
	urldate = {2025-04-04},
	publisher = {arXiv},
	author = {Kriener, Laura and Göltz, Julian and Petrovici, Mihai A.},
	month = jan,
	year = {2022},
	note = {arXiv:2102.08211 [cs]},
}

@misc{keuninckx_not_2023,
	title = {Not {Biologically} {Inspired}: {On} {Training} {Networks} of {Monostable} {Multivibrator} {Timer} {Neurons}},
	shorttitle = {({PDF}) {Not} {Biologically} {Inspired}},
	url = {https://www.researchgate.net/publication/369795246_Not_Biologically_Inspired_On_Training_Networks_of_Monostable_Multivibrator_Timer_Neurons},
	language = {en},
	urldate = {2025-04-03},
	journal = {ResearchGate},
	author = {Keuninckx, Lars and Hartmann, Matthias and Detterer, Paul and Safa, Ali and Ocket, Ilja},
	month = apr,
	year = {2023},
	doi = {10.13140/RG.2.2.27417.70242},
}

@misc{wang_lutnet_2019,
	title = {{LUTNet}: {Rethinking} {Inference} in {FPGA} {Soft} {Logic}},
	shorttitle = {{LUTNet}},
	url = {http://arxiv.org/abs/1904.00938},
	doi = {10.48550/arXiv.1904.00938},
	urldate = {2024-11-23},
	publisher = {arXiv},
	author = {Wang, Erwei and Davis, James J. and Cheung, Peter Y. K. and Constantinides, George A.},
	month = apr,
	year = {2019},
	note = {arXiv:1904.00938},
}

@misc{wang_lutnet_2020,
	title = {{LUTNet}: {Learning} {FPGA} {Configurations} for {Highly} {Efficient} {Neural} {Network} {Inference}},
	shorttitle = {{LUTNet}},
	url = {http://arxiv.org/abs/1910.12625},
	doi = {10.48550/arXiv.1910.12625},
	urldate = {2024-11-23},
	publisher = {arXiv},
	author = {Wang, Erwei and Davis, James J. and Cheung, Peter Y. K. and Constantinides, George A.},
	month = mar,
	year = {2020},
	note = {arXiv:1910.12625},
}

@article{petersen_convolutional_2024,
	title = {Convolutional {Differentiable} {Logic} {Gate} {Networks}},
	volume = {37},
	url = {https://proceedings.neurips.cc/paper_files/paper/2024/hash/db988b089d8d97d0f159c15ed0be6a71-Abstract-Conference.html},
	language = {en},
	urldate = {2025-04-03},
	journal = {Advances in Neural Information Processing Systems},
	author = {Petersen, Felix and Kuehne, Hilde and Borgelt, Christian and Welzel, Julian and Ermon, Stefano},
	month = dec,
	year = {2024},
	pages = {121185--121203},
}

@article{petersen_deep_2022,
	title = {Deep {Differentiable} {Logic} {Gate} {Networks}},
	volume = {35},
	url = {https://proceedings.neurips.cc/paper_files/paper/2022/hash/0d3496dd0cec77a999c98d35003203ca-Abstract-Conference.html},
	language = {en},
	urldate = {2025-04-03},
	journal = {Advances in Neural Information Processing Systems},
	author = {Petersen, Felix and Borgelt, Christian and Kuehne, Hilde and Deussen, Oliver},
	month = dec,
	year = {2022},
	pages = {2006--2018},
}

@article{lee_stochastic_2024,
	title = {Stochastic {Computing} {Convolutional} {Neural} {Network} {Architecture} {Reinvented} for {Highly} {Efficient} {Artificial} {Intelligence} {Workload} on {Field}-{Programmable} {Gate} {Array}},
	volume = {7},
	url = {https://spj.science.org/doi/10.34133/research.0307},
	doi = {10.34133/research.0307},
	urldate = {2025-04-03},
	journal = {Research},
	author = {Lee, Yang Yang and Halim, Zaini Abdul and Wahab, Mohd Nadhir Ab and Almohamad, Tarik Adnan},
	month = mar,
	year = {2024},
	note = {Publisher: American Association for the Advancement of Science},
	pages = {0307},
}

@inproceedings{putra_q-spinn_2021,
	title = {Q-{SpiNN}: {A} {Framework} for {Quantizing} {Spiking} {Neural} {Networks}},
	shorttitle = {Q-{SpiNN}},
	url = {https://ieeexplore.ieee.org/abstract/document/9534087?casa_token=4WX-acnbmR8AAAAA:-1TtplKY-1ovJZxSE4R5qCZutSvhYSu5sNDo8sKGN6X1Kmwl6f04EL8p4pg70EQZIhZDVxxM},
	doi = {10.1109/IJCNN52387.2021.9534087},
	urldate = {2025-04-03},
	booktitle = {2021 {International} {Joint} {Conference} on {Neural} {Networks} ({IJCNN})},
	author = {Putra, Rachmad Vidya Wicaksana and Shafique, Muhammad},
	month = jul,
	year = {2021},
	note = {ISSN: 2161-4407},
	pages = {1--8},
}

@article{rathi_stdp-based_2019,
	title = {{STDP}-{Based} {Pruning} of {Connections} and {Weight} {Quantization} in {Spiking} {Neural} {Networks} for {Energy}-{Efficient} {Recognition}},
	volume = {38},
	issn = {1937-4151},
	url = {https://ieeexplore.ieee.org/document/8325325},
	doi = {10.1109/TCAD.2018.2819366},
	number = {4},
	urldate = {2025-04-03},
	journal = {IEEE Transactions on Computer-Aided Design of Integrated Circuits and Systems},
	author = {Rathi, Nitin and Panda, Priyadarshini and Roy, Kaushik},
	month = apr,
	year = {2019},
	note = {Conference Name: IEEE Transactions on Computer-Aided Design of Integrated Circuits and Systems},
	pages = {668--677},
}

@article{eshraghian_training_2023,
	title = {Training {Spiking} {Neural} {Networks} {Using} {Lessons} {From} {Deep} {Learning}},
	volume = {111},
	issn = {1558-2256},
	url = {https://ieeexplore.ieee.org/document/10242251},
	doi = {10.1109/JPROC.2023.3308088},
	number = {9},
	urldate = {2025-04-03},
	journal = {Proceedings of the IEEE},
	author = {Eshraghian, Jason K. and Ward, Max and Neftci, Emre O. and Wang, Xinxin and Lenz, Gregor and Dwivedi, Girish and Bennamoun, Mohammed and Jeong, Doo Seok and Lu, Wei D.},
	month = sep,
	year = {2023},
	note = {Conference Name: Proceedings of the IEEE},
	pages = {1016--1054},
}

@inproceedings{rastegari_xnor-net_2016,
	address = {Cham},
	title = {{XNOR}-{Net}: {ImageNet} {Classification} {Using} {Binary} {Convolutional} {Neural} {Networks}},
	isbn = {978-3-319-46493-0},
	shorttitle = {{XNOR}-{Net}},
	doi = {10.1007/978-3-319-46493-0_32},
	language = {en},
	booktitle = {Computer {Vision} – {ECCV} 2016},
	publisher = {Springer International Publishing},
	author = {Rastegari, Mohammad and Ordonez, Vicente and Redmon, Joseph and Farhadi, Ali},
	editor = {Leibe, Bastian and Matas, Jiri and Sebe, Nicu and Welling, Max},
	year = {2016},
	pages = {525--542},
}

@article{hubara_quantized_2018,
	title = {Quantized {Neural} {Networks}: {Training} {Neural} {Networks} with {Low} {Precision} {Weights} and {Activations}},
	volume = {18},
	issn = {1533-7928},
	shorttitle = {Quantized {Neural} {Networks}},
	url = {http://jmlr.org/papers/v18/16-456.html},
	number = {187},
	urldate = {2025-04-03},
	journal = {Journal of Machine Learning Research},
	author = {Hubara, Itay and Courbariaux, Matthieu and Soudry, Daniel and El-Yaniv, Ran and Bengio, Yoshua},
	year = {2018},
	pages = {1--30},
}

@article{desislavov_trends_2023,
	title = {Trends in {AI} inference energy consumption: {Beyond} the performance-vs-parameter laws of deep learning},
	volume = {38},
	issn = {2210-5379},
	shorttitle = {Trends in {AI} inference energy consumption},
	url = {https://www.sciencedirect.com/science/article/pii/S2210537923000124},
	doi = {10.1016/j.suscom.2023.100857},
	urldate = {2025-04-03},
	journal = {Sustainable Computing: Informatics and Systems},
	author = {Desislavov, Radosvet and Martínez-Plumed, Fernando and Hernández-Orallo, José},
	month = apr,
	year = {2023},
	pages = {100857},
}

@article{schuman_opportunities_2022,
	title = {Opportunities for neuromorphic computing algorithms and applications},
	volume = {2},
	copyright = {2022 Springer Nature America, Inc.},
	doi = {10.1038/s43588-021-00184-y},
	number = {1},
	journal = {Nature Computational Science},
	author = {Schuman, Catherine D. and Kulkarni, Shruti R. and Parsa, Maryam and Mitchell, J. Parker and Date, Prasanna and Kay, Bill},
	month = jan,
	year = {2022},
	pages = {10--19},
}

@inproceedings{sevilla_compute_2022,
	title = {Compute {Trends} {Across} {Three} {Eras} of {Machine} {Learning}},
	url = {https://ieeexplore.ieee.org/document/9891914},
	doi = {10.1109/IJCNN55064.2022.9891914},
	booktitle = {2022 {International} {Joint} {Conference} on {Neural} {Networks} ({IJCNN})},
	author = {Sevilla, Jaime and Heim, Lennart and Ho, Anson and Besiroglu, Tamay and Hobbhahn, Marius and Villalobos, Pablo},
	month = jul,
	year = {2022},
	note = {ISSN: 2161-4407},
	pages = {1--8},
}

\end{document}
\typeout{get arXiv to do 4 passes: Label(s) may have changed. Rerun}